\documentclass[sigconf]{acmart}
\AtBeginDocument{%
  }

\copyrightyear{2026}
\acmYear{2026}
\makeatletter
\gdef\@copyrightpermission{
  \begin{minipage}{0.3\columnwidth}
   \href{https://creativecommons.org/licenses/by/4.0/}{\includegraphics[width=0.90\textwidth]{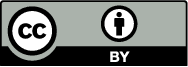}}
  \end{minipage}\hfill
  \begin{minipage}{0.7\columnwidth}
   \href{https://creativecommons.org/licenses/by/4.0/}{This work is licensed under a Creative Commons Attribution International 4.0 License.}
  \end{minipage}
  \vspace{0pt}
}
\makeatother
\acmConference[SIGIR '26]{Proceedings of the 49th International ACM SIGIR Conference on Research and Development in Information Retrieval}{July 20--24, 2026}{Melbourne, VIC, Australia}
\acmBooktitle{Proceedings of the 49th International ACM SIGIR Conference on Research and Development in Information Retrieval (SIGIR '26), July 20--24, 2026, Melbourne, VIC, Australia}
\acmDOI{10.1145/3805712.3809747}
\acmISBN{979-8-4007-2599-9/2026/07}

\usepackage{hyperref}
\usepackage{url}
\usepackage{booktabs}
\usepackage{adjustbox}
\usepackage{multicol}
\usepackage{multirow}
\usepackage{xspace}
\usepackage{algorithm}
\usepackage{algpseudocode}

\usepackage{xcolor}         % colors
\usepackage{enumitem}
\usepackage{listings}

\newcommand{\ourmethod}{\texttt{VAC}\xspace}

\begin{document}

%%
%% The "title" command has an optional parameter,
%% allowing the author to define a "short title" to be used in page headers.
\title{Learning from Natural Language Feedback for Personalized Question Answering}

%%
%% The "author" command and its associated commands are used to define
%% the authors and their affiliations.
%% Of note is the shared affiliation of the first two authors, and the
%% "authornote" and "authornotemark" commands
%% used to denote shared contribution to the research.
\author{Alireza Salemi}
\affiliation{%
  \institution{Center for Intelligent Information Retrieval}
  \institution{University of Massachusetts Amherst}
  \city{Amherst}
  \state{MA}
  \country{USA}
}
\email{asalemi@cs.umass.edu}

\author{Hamed Zamani}
\affiliation{%
  \institution{Center for Intelligent Information Retrieval}
  \institution{University of Massachusetts Amherst}
  \city{Amherst}
  \state{MA}
  \country{USA}
}
\email{zamani@cs.umass.edu}

%%
%% By default, the full list of authors will be used in the page
%% headers. Often, this list is too long, and will overlap
%% other information printed in the page headers. This command allows
%% the author to define a more concise list
%% of authors' names for this purpose.
\renewcommand{\shortauthors}{Salemi and Zamani}

%%
%% The abstract is a short summary of the work to be presented in the
%% article.
\begin{abstract}
Personalization is crucial for enhancing both the effectiveness and user satisfaction of language technologies, particularly in information-seeking tasks like question answering. Current approaches for personalizing large language models (LLMs) often rely on retrieval-augmented generation (RAG), followed by reinforcement learning with scalar reward signals to teach models how to use retrieved personal context. We believe that these scalar rewards sometimes provide weak, non-instructive feedback, limiting learning efficiency and personalization quality. We introduce \ourmethod, a novel framework for personalized response generation that replaces scalar rewards with natural language feedback (NLF) that are generated conditioned on the user profiles and the question narratives. NLF serves as a rich and actionable supervision signal, allowing the policy model to iteratively refine its outputs and internalize effective personalization strategies. Training alternates between optimizing the feedback model and fine-tuning the policy model on the improved responses, resulting in a policy model that no longer requires feedback at inference. Evaluation on the LaMP-QA benchmark that consists of three diverse domains demonstrates consistent and significant improvements over the state-of-the-art results. Human evaluations further confirm the superior quality of the generated responses. These results demonstrate that NLF provides more effective signals for optimizing personalized question answering.
\end{abstract}



% %%
% %% Keywords. The author(s) should pick words that accurately describe
% %% the work being presented. Separate the keywords with commas.
% %% A "teaser" image appears between the author and affiliation
% %% information and the body of the document, and typically spans the
% %% page.
% \begin{teaserfigure}
%   \includegraphics[width=\textwidth]{sampleteaser}
%   \caption{Seattle Mariners at Spring Training, 2010.}
%   \Description{Enjoying the baseball game from the third-base
%   seats. Ichiro Suzuki preparing to bat.}
%   \label{fig:teaser}
% \end{teaserfigure}

% \received{20 February 2007}
% \received[revised]{12 March 2009}
% \received[accepted]{5 June 2009}

%%
%% This command processes the author and affiliation and title
%% information and builds the first part of the formatted document.
\maketitle

\section{Introduction}
\label{sec:introduction}

Personalization has become a critical component in human-centered systems such as search \cite{10.1145/1462198.1462203, 10.1145/1242572.1242651, 10.1145/3539618.3591900, 10.1162/dint_a_00104}, recommendation \cite{naumov2019deep, lyu-etal-2024-llm, mao2024analysisdesignpersonalizedrecommendation}, and text generation \cite{lamp, rspg, longlamp, lampqa}, as it enhances user satisfaction, increases engagement, and improves overall system efficiency \cite{xu-etal-2025-personalized}. By tailoring outputs to individual user preferences and contexts, personalized systems can deliver more relevant and effective interactions \cite{salemi2025reasoningenhancedselftraininglongformpersonalized, longlamp}. Previous work on personalized text generation has primarily focused on content generation \cite{lamp, longlamp, salemi2025reasoningenhancedselftraininglongformpersonalized, peft-rag-personalization}, which is fundamentally different from information seeking. In content generation, the objective is to mimic the user’s writing style and preferences, whereas in information seeking, the primary goal is to deliver personalized relevant information to the user. In the context of information seeking, e.g., question answering, personalization is particularly valuable, as it enables generation of responses that align with the user's intent, background, and preferences, resulting in more accurate, relevant, and user-specific responses \cite{lampqa}.

Previous work on personalizing LLMs has mainly relied on retrieval-augmented generation (RAG) \cite{lamp, lampqa}, where personalized information is retrieved from a user profile and appended to the prompt to guide the model's output. To optimize this RAG pipeline for personalization, various approaches have been explored. For instance, when ground-truth outputs are available for a user, the model can be trained to generate these outputs conditioned on the retrieved personalized context \cite{lamp}. However, such labeled data is often unavailable across many users and tasks. To address this, reinforcement learning has been employed to further enhance the model's ability to incorporate personalized information, typically by using personalized scalar rewards---learned through a personalized reward model or derived from user-provided explanations or rubrics---that reflect the quality of generated responses \cite{salemi2025reasoningenhancedselftraininglongformpersonalized}.

These optimization methods face several limitations. First, the ground-truth output provided for a user represents only one of many potentially acceptable responses, making supervised training on a single target suboptimal and prone to local minima \cite{salemi2025reasoningenhancedselftraininglongformpersonalized}. Additionally, in the case of reinforcement learning, scalar rewards provide relatively weak supervision---they indicate whether an output is good or bad but lack actionable feedback on how to improve. Consequently, the model must infer effective adjustments without explicit guidance. Moreover, optimizing with scalar rewards often requires exploration across a wide range of outputs, leading to slow convergence and increased training cost.

\begin{figure*}
    \centering
    \includegraphics[width=\textwidth]{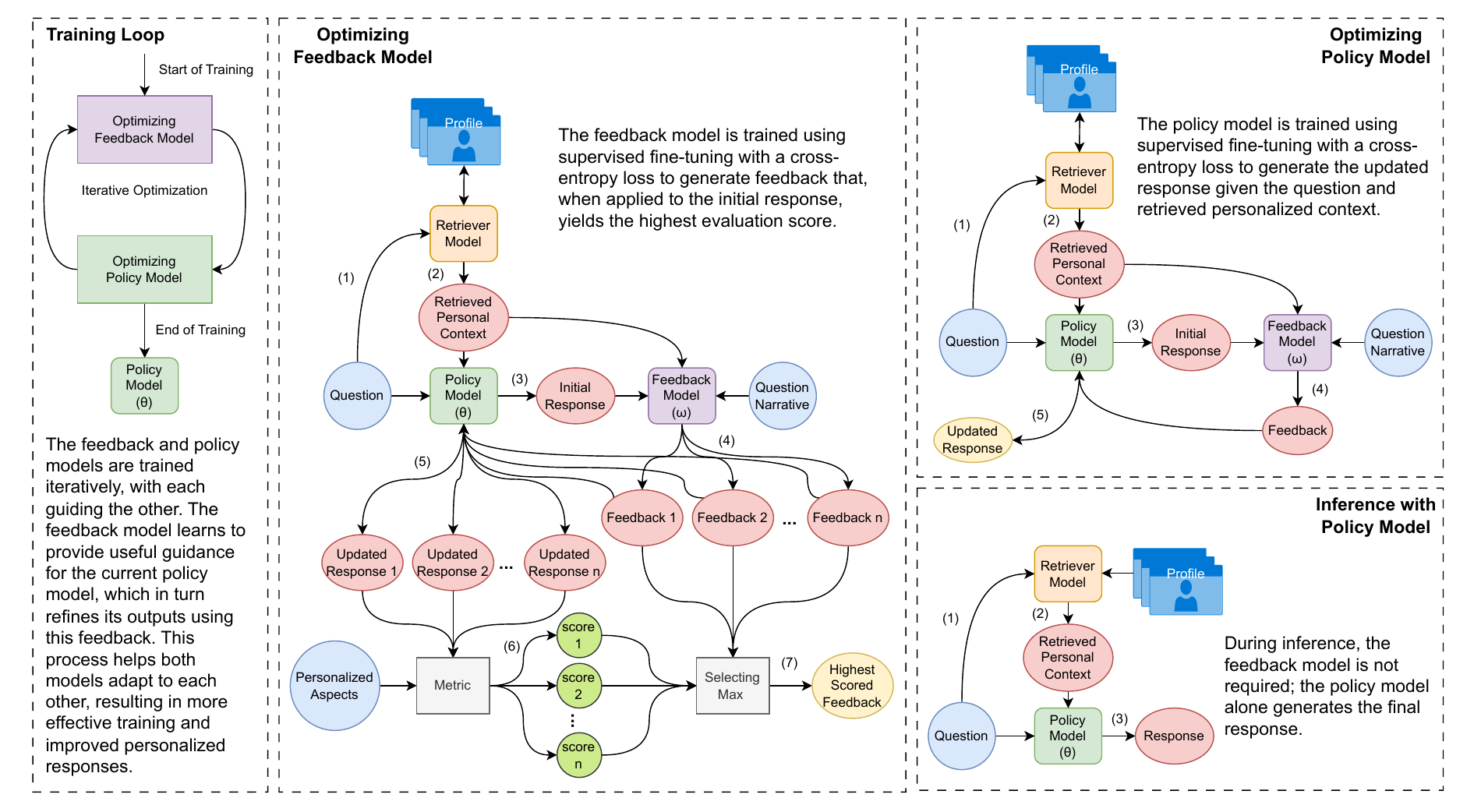}
    \vspace{-0.6cm}
    \caption{Overview of the training loop and inference process in the \ourmethod framework, illustrating the interaction between the feedback model and policy model during training, and the use of the policy model at inference time.}
    \label{fig:overview}
\end{figure*}

To address these challenges, we introduce \ourmethod,\footnote{In Sanskrit, Vāc is the personified goddess of speech, language, and wisdom. VAC also stands for Verbal-Alignment for Customization (or personalization).} a framework that replaces the scalar reward model with a feedback model that generates natural language feedback (NLF) on the policy model's responses during training. The NLF is generated based on personalized user preferences derived from the user profile and a user-authored question narrative. This feedback provides a richer and more interpretable form of supervision, guiding the policy model toward producing more personalized outputs. Training in \ourmethod, as depicted in Figure~\ref{fig:overview}, follows an iterative process over several rounds. In each iteration, the feedback model is first optimized to generate feedback that, when used to revise a response by the current policy model, leads to measurable improvements according to a task-specific evaluation metric. Once trained, the feedback model generates feedback for the responses produced by the current policy model. This feedback is then used to guide the policy model in editing its initial responses to improve them for the user. Finally, the policy model is fine-tuned via supervised learning to generate the improved responses directly from the input, eliminating the need for feedback at inference time. Compared to scalar rewards, which offer only coarse and indirect supervision signals, natural language feedback provides explicit and actionable guidance. Training the feedback model on outputs generated by the current policy model allows it to adapt to the model's evolving behavior and capabilities, resulting in more targeted and effective feedback. Conversely, training the policy model on feedback-refined responses enables it to internalize effective personalization patterns, improving its generation quality without relying on feedback at test time.

To evaluate \ourmethod, we conduct experiments on the recent Language Model Personalized Question Answering benchmark (LaMP-QA) \cite{lampqa}, which includes three diverse domains. Our results demonstrate that \ourmethod consistently outperforms all baselines, achieving a 13.6\% relative improvement over the non-personalized baseline, a 3.6\% improvement over the best-performing personalized baseline while being 1.9$\times$ more efficient in terms of inference time, and a 6.0\% improvement over reinforcement learning with scalar rewards. Additionally, human evaluation shows that \ourmethod is preferred in 44\% of the cases, ties in 33\%, and is less preferred in only 23\% of comparisons against the state-of-the-art method. Furthermore, we provide detailed ablation studies analyzing various aspects of the proposed method, including the impact of different optimization strategies and the size of the feedback model on overall performance. To facilitate future research, we publicly release our code and data.
% \footnote{The codes will be released after acceptance of the paper.}
\footnote{Available at: \url{https://github.com/alirezasalemi7/VAC}}

\section{Related Work}

\subsubsection*{\textbf{Personalization}}

Personalization plays a central role in search, recommendation, and text generation \cite{10.1145/2702123.2702503, 10.1145/1462198.1462203, naumov2019deep, lamp, longlamp, salemi2025reasoningenhancedselftraininglongformpersonalized}, as it has been shown to improve user satisfaction, efficiency, and long-term engagement \cite{lampqa, expert}. Personalization is particularly beneficial for question answering, as it enables models to generate responses that are better aligned with the user's preferences, background, and prior knowledge, ultimately leading to more relevant and effective answers \cite{lampqa}. In this work, we focus on personalized question answering, a setting for which, to the best of our knowledge, LaMP-QA \cite{lampqa} is the only publicly available benchmark.

To personalize an LLM, \citet{lamp} proposed a RAG framework that retrieves information from the user profile and incorporates it into the prompt provided to the LLM. Furthermore, \citet{salemi2025reasoningenhancedselftraininglongformpersonalized} extend this approach by optimizing the LLM with reinforcement learning to better incorporate retrieved personal context. Beyond this line of work, existing methods for personalization span a range of strategies, including training retrievers with personalized relevance feedback \citep{rspg}, fine-tuning LLMs with user-specific supervision \citep{jang2023personalized}, and designing prompts tailored to individual users \citep{Li_2024}. Parameter-efficient fine-tuning has also been explored for personalized generation \citep{tan2024personalized}, with recent efforts integrating these techniques into RAG pipelines \citep{peft-rag-personalization}. In addition, reasoning and self-training have shown promise in improving long-form personalized generation \citep{salemi2025reasoningenhancedselftraininglongformpersonalized}. Personalized assistants have been studied across a variety of domains, including education and enterprise applications \citep{li2023teach, mysore2023pearl, lu2024corporate, zhang-etal-2024-llm-based}. Despite interest in personalized generation, personalized question answering remains relatively underexplored.

\subsubsection*{\textbf{Learning from Natural Language Feedback}}

Due to the expressive nature of language, NLF has been explored as a training signal in tasks such as mathematical reasoning and code generation, where ground truth answers are well-defined \citep{chen2024learning, chen2024improvingcodegenerationtraining}. These studies demonstrate that human-written NLF can substantially improve model performance, while feedback generated by other LLMs tends to be less effective. At inference time, NLF has also been used in collaborative setups, where two models jointly solve a task through iterative feedback \citep{paul-etal-2024-refiner, li-etal-2025-learning-reason, xi2024enhancingllmreasoningcritique, yan-etal-2023-learning, yang2024large}. Another line of work leverages NLF at inference to optimize prompts rather than model parameters \citep{textgrad}, though such methods typically introduce latency at test time.

This paper provides the first attempt on using  NLF for personalization. Our work departs from priors in several key ways. First, unlike domains such as math and code where answers are either correct or incorrect, personalization requires learning subjective user preferences, where a response may be suitable for one user but not for another. Second, we automatically generate feedback conditioned on the information about the user, removing the need for human supervision during training. Third, our method operates within a RAG framework, where both the retrieval and generation components contribute to the policy and feedback models' performance. Finally, we propose a joint training procedure that alternates between optimizing the feedback and policy model, allowing them to co-adapt and resulting in more effective learning.

\section{Problem Formulation}

We consider a setting in which a user $u$ is associated with a profile $P_u = \{d_i^u\}_{i=1}^{|P_u|}$, consisting of their previous questions and corresponding detailed descriptions. Given a new query $x_u$, an LLM $\pi_{\theta}$ generates a personalized response $y_{x_u} = \pi_{\theta}(P_u, x_u)$ by conditioning on both $P_u$ and $x_u$. To evaluate the quality of the generated response, we assume access to a set of $n_{x_u}$ user-specific aspects $E_{x_u} = \{e_i\}_{i=1}^{n_{x_u}}$, which are extracted from a personalized question narrative $r_{x_u}$ provided by the user. These aspects are used exclusively for evaluation and are not accessible to the policy model during generation. A metric $\mu(x_u, \hat{y}_u, E_{x_u}, r_{x_u})$ quantifies the quality of the response based on the extent to which the expected aspects are addressed in the generated output. Since the aspects are explicitly derived from user-provided requirements, this evaluation framework enables a targeted assessment of how well the response aligns with the user's personalized information needs. Finally, we assume the existence of a training dataset $D = \{(x_i, P_i, E_{x_i}, r_{x_i})\}_{i=1}^{|D|}$ that our method can learn from. In the test set, the structure is identical, but the aspects and narratives are used exclusively for evaluation and are never provided to the policy model.

\section{The \ourmethod Framework}

As discussed in Section~\ref{sec:introduction}, optimizing the policy model $\pi_{\theta}$ with scalar rewards is limited in effectiveness, as these rewards offer only coarse feedback indicating overall output quality, without specifying how the output should be improved. Consequently, the model must explore and infer appropriate adjustments on its own, which slows convergence and increases the cost of training. 

To address these challenges, we proposes the \ourmethod framework which replaces the scalar reward model with a feedback model $\phi_\omega$ that generates natural language feedback (NLF) on the outputs of the policy model to guide it toward more personalized responses during training. The framework follows an iterative training procedure in which the feedback and policy models are alternately optimized over multiple rounds. This enables the feedback model to improve its ability to generate effective, personalization-oriented feedback to guide the policy model to generate more personalized responses, while the policy model progressively learns to produce more personalized responses without relying on feedback during inference. The remainder of this section details our proposed method.

\begin{figure}
    \centering
    % \vspace{-0.4cm}    
    \includegraphics[width=\linewidth]{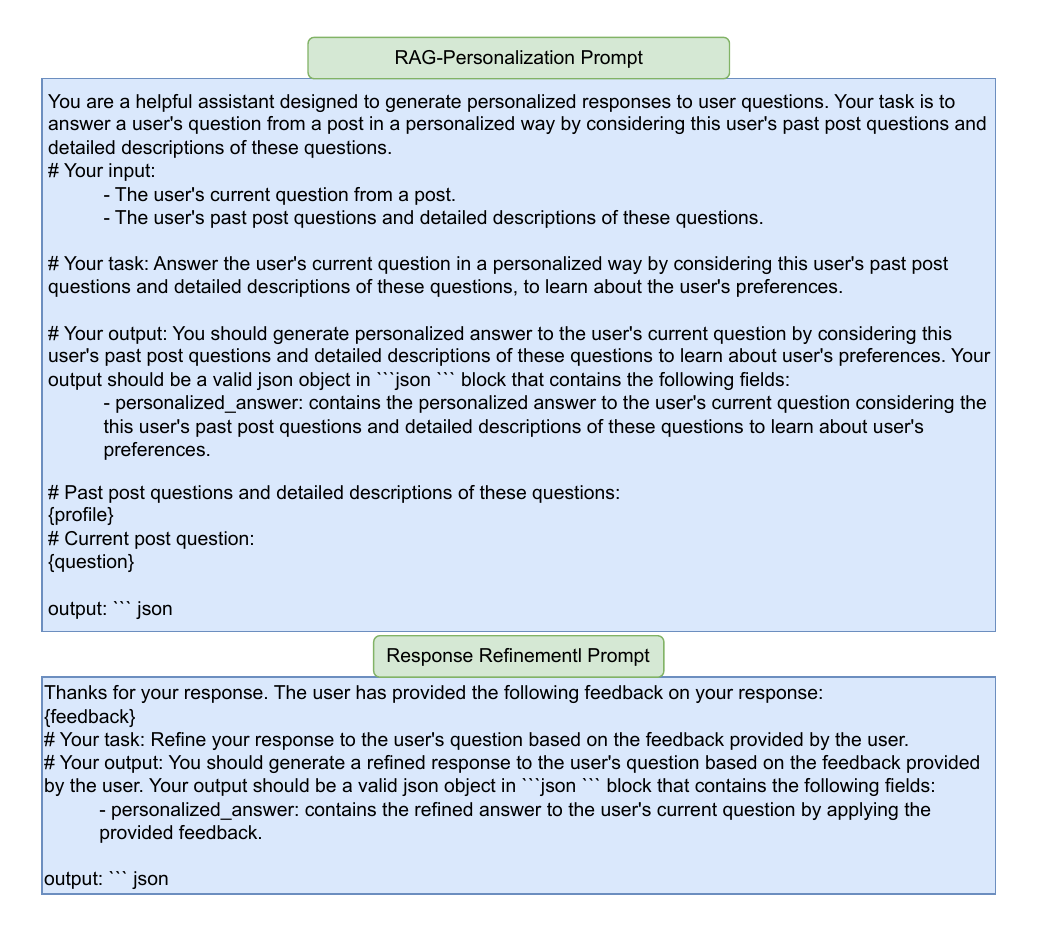}
    \vspace{-0.8cm}
    \caption{Prompts used for response generation in RAG-Personalization, Offline RL Personalization, and \ourmethod (top), and for response refinement in \ourmethod (bottom).}
    \label{fig:rag-personalization-prompt}
    % \vspace{-0.3cm}
\end{figure}

\subsubsection*{\textbf{Overview of the Training Pipeline:}}

The overview of the \ourmethod's iterative training loop is illustrated in Figure~\ref{fig:overview} and Algorithm~\ref{alg:our-method}. Over $T$ iterations, each iteration $t$ begins by optimizing the feedback model $\phi_{\omega^t}$ using offline reinforcement learning to learn an effective feedback strategy that improves the personalized responses of the policy model $\pi_{{\theta}^{t-1}}$, based on the response distribution generated by $\pi_{{\theta}^{t-1}}$. This optimization process can be formalized as:
\begin{equation}
    \omega^{t} = \underset{\omega}{\mathrm{argmax}} \frac{1}{|D|}\sum_{d \in D} U_{\text{feedback}}(\phi_{\omega};d; \pi_{\theta^{t-1}})
\end{equation}
where $D$ is the training dataset and $U_{\text{feedback}}$ is a objective function that measures the utilization of the feedback generated by the feedback model for the policy model $\pi_{\theta^{t-1}}$ based on the how it improves the evaluation metric $\mu$, as will be described in Section~\ref{sec:optimization}.

Once the feedback model trained, in the same iteration, the feedback model $\phi_{\omega^t}$ is used to produce feedback for outputs generated by $\pi_{{\theta}^{t-1}}$, and the policy model is asked to revise its responses accordingly. The updated responses are then used to fine-tune a new policy model $\pi_{{\theta}^{t}}$ using supervised learning, allowing it to better personalize its responses for users in subsequent iterations without relying on feedback at inference time. This is formalized as:
\begin{equation}
    \theta^{t} = \underset{\theta}{\mathrm{argmax}} \frac{1}{|D|}\sum_{d \in D} U_{\text{policy}(\pi_{\theta};d;\phi_{\omega^t})}
\end{equation}
where $D$ is the training dataset and $U_{\text{policy}}$ is a objective function for the policy model that encourages the model using supervised learning to generated the updated response after applying feedback from feedback model $\phi_{\omega^t}$ without relying on feedback during inference, as will be described in Section~\ref{sec:optimization}.

The details of the feedback generation and response refinement process are provided in Section~\ref{sec:feedback-refinement} and the optimization procedures for both the policy and feedback models are described in Section~\ref{sec:optimization}.

\subsubsection*{\textbf{Overview of the Inference Pipeline:}}

For inference, as shown in Figure~\ref{fig:overview} (Inference w/ policy model), we adopt the same RAG pipeline as \citet{lampqa}, where for a question $x_u$ from user $u$ with profile $P_u$, a retriever $R$ selects $K$ relevant documents from $P_u$. These retrieved documents are then appended to the question to form the prompt and fed into the trained policy model $\pi_{\theta^T}$ to generate the response using the prompt shown in Figure~\ref{fig:rag-personalization-prompt} (top), formalized as $y = \pi_{\theta^T}(x_u; R(x_u, P_u, K))$.

\begin{algorithm*}
\caption{Implementation of the training loop in the \ourmethod framework.}\label{alg:our-method}
\begin{algorithmic}[1]
\Require policy model $\pi_{{\theta}^{0}}$, feedback model $\phi_{\omega^0}$, retriever $R$, dataset $D$, metric $\mu$, number of training iterations $T$, number of retrieved documents $K$, number of generated feedback $N$
\Ensure trained policy model $\pi_{\theta^{T}}$, trained feedback model $\phi_{\omega^T}$

\For{$t = 1$ until $T$}
\State {// training the feedback model $\phi_{\omega^t}$ for round $t$}
\State $D_{\phi_{\omega^t}} = \{\}$ \Comment{This round's training data for feedback model}
\For{$(x_u, P_u, E_{x_u}, r_{x_u}) \in D_{\text{train}}$} \Comment{For each input in training dataset}
\State $y_{t-1} = \pi_{{\theta}^{t-1}}(x_u; R(x_u, P_u, K))$ \Comment{Generate initial output}
\State $F_{} = \{\}$ \Comment{Set of feedbacks for this specific output}
\For{$j = 1$ until $N$} \Comment{For $N$ times}
\State $F = F \cup \{\phi_{\omega^{t-1}}(x_u; R(x_u, P_u, K); r_{x_u}; y_{t-1})\}$ \Comment{Generate a feedback using feedback model with a high temperature for the generated output}
\EndFor
\State $D_{\phi_{\omega^t}} = D_{\phi_{\omega^t}} \cup \{(x_u, y_{t-1}, P_u, r_{x_u}, f) | \underset{f \in F}{\mathrm{argmax}}\, \mu (x_u, \pi_{{\theta}^{t-1}}(x_u; R(x_u, P_u, K); y_{t-1}; f), E_{x_u}, r_{x_u})\}$
\Comment{Find the feedback that maximizes the metric when applied to the previous generated output and add it to the training set}
\EndFor
\State $\omega^{t} = \underset{\omega}{\mathrm{argmax}}\, \sum_{(x_u, y_{t-1}, r_{x_u}, f) \in D_{\phi_{\omega^t}}} \log p_{\omega}(f | x_u; R(x_u, P_u, K); r_{x_u}; y_{t-1})$ \Comment{Maximize the probability of generating feedback given the generated output, input, and personalized aspects}
\State {// training the policy model $\pi_{{\theta}^{t}}$ for round $t$}
\State $D_{\pi_{\theta^t}} = \{\}$ \Comment{This round's training data for policy model}
\For{$(x_u, P_u, E_{x_u}, r_{x_u}) \in D_{\text{train}}$} \Comment{For each input in training dataset}
\State $y_{t-1} = \pi_{{\theta}^{t-1}}(x_u; R(x_u, P_u, K))$ \Comment{Generate initial output}
\State $f = \phi_{\omega^t}(x_u; R(x_u, P_u, K); r_{x_u}; y_{t-1})$ \Comment{Generate a feedback using the optimized feedback model}
\State $y_{t} = \pi_{{\theta}^{t-1}}(x_u; R(x_u, P_u, K); y_{t-1}; f)$ \Comment{Apply the generated feedback to the previous generated output}
\State $D_{\pi_{\theta^t}} = D_{\pi_{\theta^t}} \cup \{(x_u, P_u, y_{t})\}$ \Comment{Add the updated output to the training set}
\EndFor
\State ${{\theta}^{t}} = \underset{\theta}{\mathrm{argmax}}\, \sum_{(x_u, P_u, y_{t}) \in D_{\pi_{\theta^t}}} \log p_{\theta}(y_{t}|x_u; R(x_u, P_u, K))$ \Comment{Maximize the probability of generating the updated output given the inputs and retrieved personal documents}
\EndFor
\State \Return{$\pi_{\theta^T}, \phi_{\omega^T}$} \Comment{Return the fully trained policy and feedback model}
\end{algorithmic}
\end{algorithm*}

\subsection{Feedback Generation \& Output Refinement}
\label{sec:feedback-refinement}

Given an initial personalized response $y_{t-1} = \pi_{{\theta}^{t-1}}(x_u;R(x_u, P_u, K))$ to a question $x_u$ for user $u$ in iteration $t$ generated by the previous iteration policy model $\pi_{{\theta}^{t-1}}$, this section outlines the procedure for generating NLF and refining the response accordingly.

\subsubsection*{\textbf{Feedback Generation}}

\begin{figure}
    \centering
    \includegraphics[width=\linewidth]{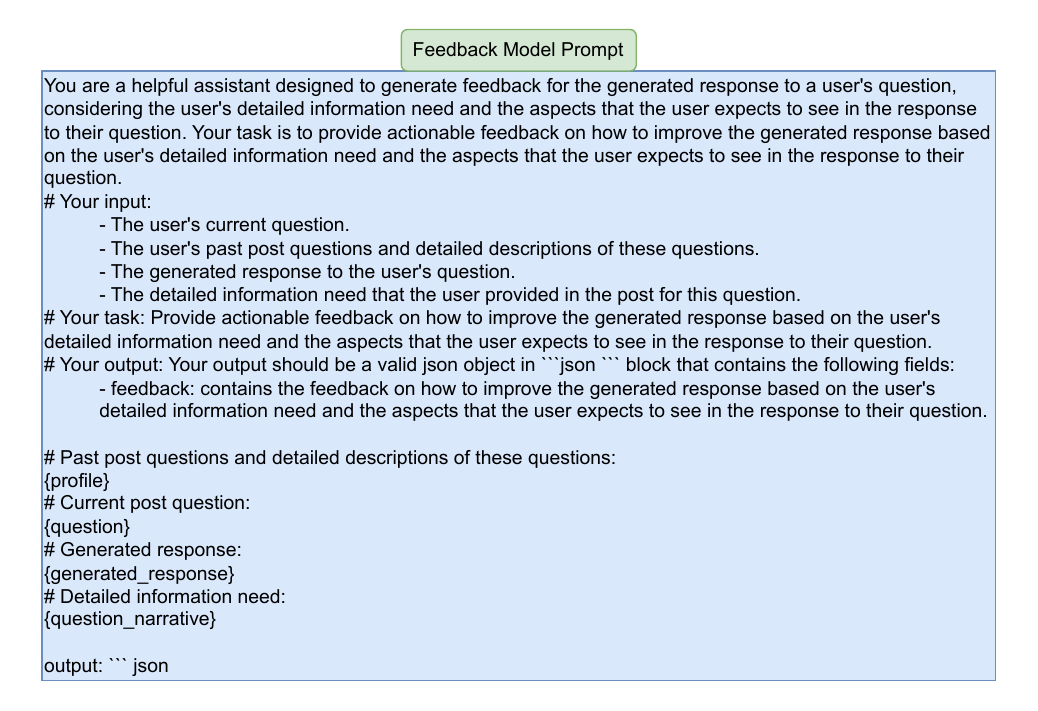}
    \vspace{-0.8cm}
    \caption{Prompt used with the feedback model in the \ourmethod framework to generate NLF on the initial output.}
    \label{fig:feedback-prompt}
    % \vspace{-0.5cm}
\end{figure}

To generate feedback for the initial personalized response $y_{t-1}$ in iteration $t$, we first retrieve $K$ documents from the user profile $P_u$ using the retriever $R$, conditioned on the question $x_u$. Then, given the question, the retrieved documents, the initial response, and the question narrative $r_{x_u}$ for the user $u$, we prompt the feedback model $\phi_{\omega^t}$ to analyze the response and produce NLF aimed at improving personalization of the response, guided by the narrative. The prompt used for this process is illustrated in Figure~\ref{fig:feedback-prompt}. Formally, the feedback is defined as: $f = \phi_{\omega^t}(x_u; R(x_u, P_u, K); r_{x_u}; y_{t-1})$.

\subsubsection*{\textbf{Output Refinement}}

To refine the initial output of iteration $t$ denoted as $y_{t-1}$ using feedback $f$, we append the initially retrieved documents from the user profile with the initial response $y_{t-1}$ in iteration $t$ and the feedback, and prompt the policy model $\pi_{{\theta}^{t-1}}$ to revise its response based on this information. This process is illustrated in the prompt shown in Figure~\ref{fig:rag-personalization-prompt} (bottom) and is formally defined as: $y_{t} = \pi_{{\theta}^{t-1}}(x_u;R(x_u, P_u, K); y_{t-1}; f)$.

\subsection{Feedback \& Policy Model Optimization}
\label{sec:optimization}

This section describes training objectives for the feedback and policy model, which are optimized iteratively as shown in Algorithm~\ref{alg:our-method}.

\subsubsection*{\textbf{Optimizing the Feedback Model}}

This section defines the $U_{\text{feedback}}$ objective function. The primary goal of the feedback model $\phi_{\omega^t}$ at iteration $t$ is to generate feedback that effectively improves the personalized responses produced by the previous policy model $\pi_{{\theta}^{t-1}}$, so that the improved responses can provide useful supervision for training the next policy model $\pi_{{\theta}^{t}}$. Accordingly, the optimization objective of the feedback model should be designed to encourage the generation of feedback that, when used by the policy model to revise its initial output, leads to a measurable improvement in the quality and personalization of the response.

To optimize the feedback model $\phi_{\omega^t}$, we follow a self-training approach as outlined in Algorithm~\ref{alg:our-method} (lines 2–12). For each input example in the dataset, we first generate an initial personalized response using the policy model from the previous iteration, $\pi_{{\theta}^{t-1}}$. Then, using the previous iteration feedback model $\phi_{\omega^{t-1}}$, we generate $N$ diverse feedback candidates by sampling at a high temperature (temperature = 1). Each candidate feedback is then applied independently to the response $y_{t-1}$ using $\pi_{{\theta}^{t-1}}$ to produce a revised response. These revised responses are subsequently evaluated using the personalized rubrics provided in the dataset, based on the downstream task-specific metric (as explained in Section~\ref{sec:dataset}). The feedback that leads to the highest evaluation score is selected as the most effective one (line 10 in Algorithm~\ref{alg:our-method}). This selected feedback is then used as a supervision signal to train the current feedback model $\phi_{\omega^t}$ using supervised fine-tuning with a cross-entropy loss \cite{seq2seq} (line 12 in Algorithm~\ref{alg:our-method}). The model is trained to generate this effective feedback given the question, the initial response, the retrieved documents, and the question narrative. This optimization encourages $\phi_{\omega^t}$ to generate feedback that, when used to revise the policy model's output, results in measurable improvements according to the task-specific evaluation metric. In this way, the objective function $U_\text{feedback}$ is formulated such that optimizing it results in feedback that is effective in improving the policy model’s outputs.

\subsubsection*{\textbf{Optimizing the Policy Model}}

This section defines the $U_{\text{policy}}$ objective function. As shown in Algorithm~\ref{alg:our-method} (lines 13–21), the optimization of the policy model proceeds as follows: For each input in the training dataset, the policy model from the previous iteration, $\pi_{{\theta}^{t-1}}$, first generates an initial response $y_{t-1}$ to the input query. Next, the feedback model $\phi_{\omega^t}$, trained in the current iteration to align itself with the policy model $\pi_{{\theta}^{t-1}}$, produces a feedback signal for each initial response. This feedback is then used by $\pi_{{\theta}^{t-1}}$ to revise its initial output, producing an improved response (line 18 in Algorithm~\ref{alg:our-method}). Finally, the updated policy model $\pi_{{\theta}^{t}}$ is trained using supervised fine-tuning with a cross-entropy loss \cite{seq2seq}, learning to generate the improved response directly from the input query (line 21 in Algorithm~\ref{alg:our-method}). This design of the objective function $U_{\text{policy}}$ assumes that the policy model’s updated response after applying feedback is a better personalized response for the user. Consequently, the model is trained to reproduce this response directly, without requiring feedback. This training procedure helps the policy model learn to produce higher-quality, personalized responses without relying on feedback at inference time.

\section{Experiments}

\subsection{Experimental Setup}
\label{sec:dataset}

\begin{table*}
    \centering
    \caption{Dataset statistics of the each dataset in the LaMP-QA benchmark.}
    % \vspace{-0.3cm}
    \adjustbox{max width=\textwidth}{
    \begin{tabular}{l|ccc|ccc|ccc}
        \toprule
        \multirow{3}{*}{\textbf{Attribute}} & \multicolumn{3}{c}{\textbf{Arts \&}} & \multicolumn{3}{c}{\textbf{Lifestyle \& Personal}} & \multicolumn{3}{c}{\textbf{Society \&}} \\
        & \multicolumn{3}{c}{\textbf{Entertainment}} & \multicolumn{3}{c}{\textbf{Development}} & \multicolumn{3}{c}{\textbf{Culture}} \\
        \cmidrule{2-10}
        & train & validation & test & train & validation & test & train & validation & test \\
        \midrule
        \textbf{\#Questions (users)} & 9349 & 801 & 767 & 7370 & 892 & 989 & 7614 & 810 & 1074  \\
        \midrule
        \textbf{\#Evaluation Aspects} & $2.7\pm0.9$ & $4.7\pm1.2$ & $4.6\pm1.2$ & $3.1\pm1.0$ & $5.1\pm1.1$ &  $5.1\pm1.2$ & $2.9\pm0.9$ & $4.8\pm1.1$ & $4.8\pm1.0$ \\
        \midrule
        \textbf{Profile Size} & $106.7 \pm 127.3$ & $129.0 \pm 183.7$ &  $159.1 \pm 203.0$ & $116.6 \pm 162.0$ &  $98.2 \pm 198.6$ & $111.6 \pm 220.3$ &  $141.3 \pm 194.7$ & $110.5 \pm 210.6$ &  $115.8 \pm 203.6$ \\
        % \midrule
        % \textbf{Question Length} & $13.0 \pm 2.9$ & $10.6 \pm 4.0$ & $10.0 \pm 3.8$ & $13.6 \pm 3.3$ & $11.3 \pm 4.4$ & $11.6 \pm 4.6$ & $14.2 \pm 3.6$ & $12.1 \pm 4.9$ & $12.9 \pm 5.4$ \\
        % \midrule
        % \textbf{Narrative Length} & $113.1 \pm 98.2$ & $166.1 \pm 167.6$ &  $144.7 \pm 146.0$ & $132.2 \pm 104.1$ & $159.2 \pm 138.5$ & $169.4 \pm 145.2$ & $144.6 \pm 117.9$ & $161.6 \pm 158.2$ & $167.9 \pm 143.4$ \\
        \bottomrule
    \end{tabular}}
    \label{tab:dataset-stats}
\end{table*}

\subsubsection*{\textbf{Datasets \& Evaluation:}}

We conduct our experiments on the only publicly available dataset for personalized question answering, the LaMP-QA benchmark \cite{lampqa}, which comprises three diverse domains: (1) Art \& Entertainment, (2) Lifestyle \& Personal Development, and (3) Society \& Culture.
% \footnote{Benchmarks such as LaMP \cite{lamp} and LongLaMP \cite{longlamp} are not suitable for our setting because: (1) they focus on content generation tasks, which is fundamentally different from our focus on question answering, and (2) unlike LaMP-QA, which provides user expectations about the response as ground-truth labels, they offer only a single ground-truth response. This makes them less suitable for studying personalization and potentially misaligned with real user expectations, as discussed by \citet{lampqa}, and is particularly limiting for our feedback-provision method, where converging to a single response is both infeasible and suboptimal for training and evaluation.} 
\footnote{LaMP \cite{lamp} and LongLaMP \cite{longlamp} are other available datasets for personalized text generation; however, they are not well-suited for our setting for two main reasons.
First, their primary focus is on content generation tasks, which differs fundamentally from our focus on question answering as explained in Section~\ref{sec:introduction}. Second, unlike LaMP-QA---which provides explicit annotations of user expectations as ground-truth---they include only a single reference response. This design limits their ability to represent the diversity of valid personalized outputs and risks misalignment with real user expectations, as noted by \citet{lampqa}. Such limitations are especially restrictive for our feedback-provision method, where converging to a single response is both infeasible and suboptimal for robust training and fair evaluation.}
Each example in these datasets includes a user query, the user's question history serving as their profile, a question narrative that reflects the user's perspective and intent, and a set of personalized rubrics that specify the aspects that an ideal response should address. The statistics of the datasets used in our experiments are shown in Table~\ref{tab:dataset-stats}. To evaluate responses, following \citet{lampqa}, we employ the instruction-tuned Qwen 2.5 model with 32 billion parameters\footnote{Available at: \url{https://hf.co/Qwen/Qwen2.5-32B-Instruct}} \cite{qwen2.5}. For each question, the LLM assesses whether each personalized aspect is addressed in the response, assigning a score in the range $[0, 2]$. The scores are then normalized to the range $[0, 1]$. The final score for a generated response is computed as the average normalized score across all personalized aspects. For more information about this we refer the reader to \citet{lampqa}. Additionally, in one experiment, we confirm our main experimental result through side-by-side human evaluation between \ourmethod and the best baseline.

% \begin{algorithm}
% \caption{Implementation of the official evaluation metric for the LaMP-QA benchmark.}\label{alg:eval}
% \begin{algorithmic}[1]
% \Require prompt $x_u$, requirement details $r_u$, important aspects $E_{x_u}$, generated response $\hat{y}_{x_u}$, Evaluator LLM $\pi$
% \Ensure score $s_{\hat{y}_{x_u}}$
% \State $s^{t}_{\hat{y}_{x_u}} = 0$ \Comment{Score initialization with zero}
% \For{$e_i \in E_{x_u}$} \Comment{Scoring output based on each aspect}
% \State $s^{t}_{e_i} = \pi(x_u, r_u, e_i, \hat{y}_{x_u})$ \Comment{Scoring output based on aspect using prompt in Figure~\ref{fig:eval-prompt}}
% \State $s_{e_i} = \frac{s^{t}_{e_i}}{2}$ \Comment{Normalizing the aspect score for aspect by division by 2}
% \State $s^{t}_{\hat{y}_{x_u}} = s^{t}_{\hat{y}_{x_u}} + s_{e_i}$ \Comment{Score accumulation for averaging}
% \EndFor
% \State $s_{\hat{y}_{x_u}} = \frac{s^{t}_{\hat{y}_{x_u}}}{|E_{x_u}|}$ \Comment{Averaging the output score using division by the number of aspects}
% \State \Return $s_{\hat{y}_{x_u}}$ \Comment{Returning score for output for user $u$}
% \end{algorithmic}
% \end{algorithm}

% \begin{figure}[!ht]
%     \centering \includegraphics[width=\linewidth]{figs/prompt-metric.pdf}
%     \vspace{-0.8cm}
%     \caption{Evaluation prompt for assessing the quality of generated responses using the personalized aspects.}
%     \label{fig:eval-prompt}
% \end{figure}

\subsubsection*{\textbf{Training Configuration:}}

We use the instruction-tuned Qwen 2.5 model with 7 billion parameters\footnote{Available at: \url{https://hf.co/Qwen/Qwen2.5-7B-Instruct}} \cite{qwen2.5} as the policy LLM. Extending our approach to larger or alternative LLMs would require over 750 GPU hours, which is beyond our computational budget; thus, we limit our experiments to a single LLM, which constitutes a limitation of this work. Unless otherwise specified, the feedback model uses the same backbone LLM (instruction-tuned Qwen 2.5 model with 7 billion parameters). We conduct training for $T = 3$ iterations. For training feedback provider, we generate $N = 16$ feedback per output. Each iteration uses the best checkpoint from the previous iteration to initialize the models' weights. Training is performed using the Adam optimizer \citep{adam} with a learning rate of $5 \times 10^{-5}$, a batch size of 32, and gradient clipping with a maximum norm of 1 for stability. Models are trained for up to 5000 steps with a warmup phase over the first 10\% of steps, followed by linear learning rate decay. We fine-tune the models using Low-Rank Adaptation (LoRA) \cite{lora}, with rank $r = 16$, scaling factor $\alpha = 16$, and a dropout rate of 0.05, applied without modifying bias parameters. LoRA is implemented using the PEFT library.\footnote{Available at: \url{https://github.com/huggingface/peft}} Model checkpoints are evaluated every 250 steps on the validation set, and the best-performing checkpoint is selected for evaluation. All experiments are conducted using 4 NVIDIA A100 GPUs with 80GB of VRAM and 128GB of RAM, completed in around 750 GPU hours.

\begin{figure}
    % \vspace{-0.4cm}
    \centering
    \includegraphics[width=\linewidth]{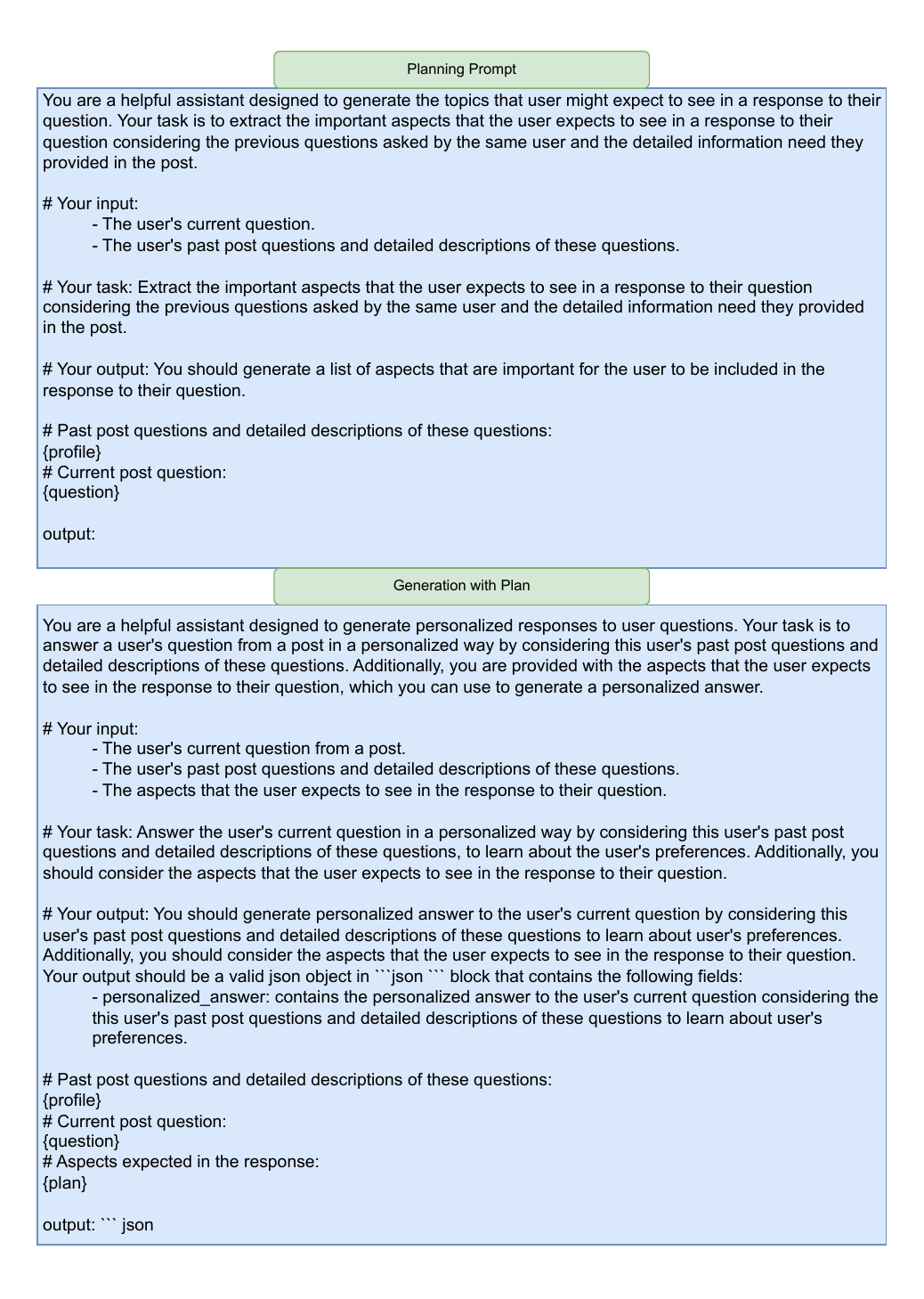}
    \vspace{-1.2cm}
    \caption{Prompt used for PlanPers \cite{lampqa} baseline.}
    \label{fig:planpers-prompt}
    % \vspace{-0.6cm}
\end{figure}

\subsubsection*{\textbf{Inference Configuration:}}

All models are configured with a maximum input-output token limit of 8192 tokens. Response generation is performed using nucleus sampling \cite{nu_sampling} with a temperature of $0.1$. To enable efficient inference and deployment of LLMs, we utilize the VLLM library.\footnote{Available at: \url{https://github.com/vllm-project/vllm}} For retrieval, we employ Contriever \cite{contriever}, a dense retriever fine-tuned on the MS MARCO dataset \cite{msmarco}, to retrieve $k = 10$ relevant documents from the user profile.

\begin{table*}
    \centering
    \caption{Performance on the test set of the LaMP-QA benchmark. The superscript $^\dagger$ shows a statistically significant difference between the best-performing baseline and our method using t-test ($p < 0.05$).}
    \vspace{-0.3cm}
    \adjustbox{max width=\linewidth}{\begin{tabular}{l|c|ccc|c}
    \toprule
    \multirow{2}{*}{\textbf{Method}} & \textbf{Runtime} & \textbf{Arts \&} & \textbf{Lifestyle \& Personal} & \textbf{Society \&} & \textbf{Average}  \\
    & (second / query) & \textbf{Entertainment} & \textbf{Development} & \textbf{Culture} & \textbf{(macro)} \\
    \midrule
    No-Personalization & 0.78 & 0.3129 & 0.4582 & 0.4769 & 0.416 \\
    RAG-Personalization (Random $P$) & 1.65 & 0.2547 & 0.3829 & 0.4037 & 0.3471 \\
    RAG-Personalization & 1.71 & 0.3397 & 0.4481 & 0.4967 & 0.4281 \\
    PlanPers & 3.12 & {0.3518} & {0.4818} & {0.5240} & {0.4525} \\
    Offline RL RAG-Personalization & 1.67 & \textbf{0.3579} & 0.4621 & 0.5070 & 0.4423 \\
    \midrule
    \textbf{\ourmethod} & 1.63 & 0.3454 & \textbf{0.5288$^\dagger$} & \textbf{0.5331$^\dagger$} & \textbf{0.4691$^\dagger$} \\
    \bottomrule
    \end{tabular}}
    \label{tab:main-results-test}
    % \vspace{-0.4cm}
\end{table*}

\begin{figure*}
    \centering
    \includegraphics[width=\textwidth]{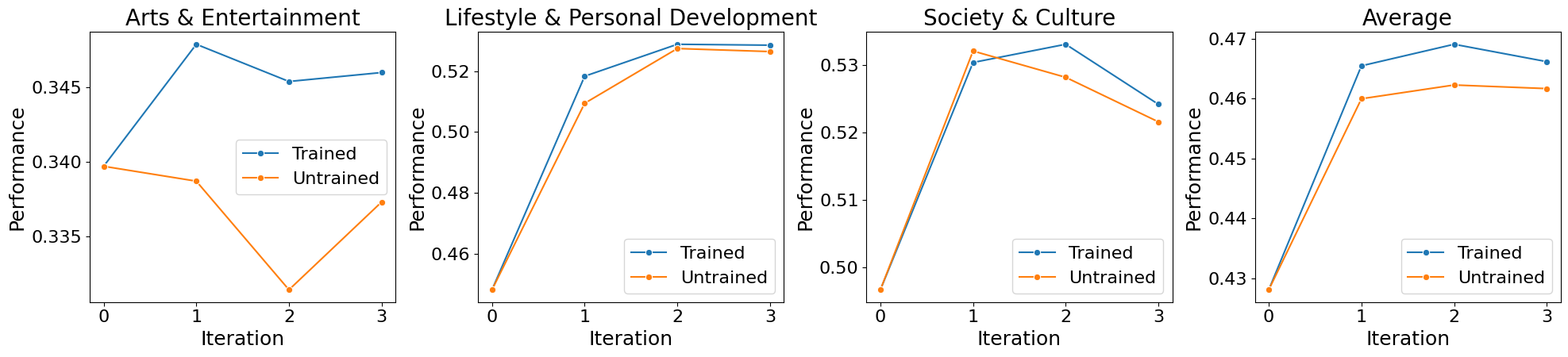}
    \vspace{-0.6cm}
    \caption{Performance of \ourmethod in different training iterations with trained and untrained feedback model.}
    \label{fig:iteration-test}
    % \vspace{-0.3cm}
\end{figure*}

\subsubsection*{\textbf{Baselines:}}

We compare \ourmethod against a set of personalized and non-personalized baselines. For the non-personalized baseline, we directly provide the question to the LLM without any user context. For personalized baselines, we include the following:
\begin{itemize}[leftmargin=1em]
    \item RAG-Personalization \cite{lamp, lampqa}: The question is used to retrieve relevant documents from the user profile, which then the LLM generates a response using both the query and the retrieved personal context with the prompt shown in Figure~\ref{fig:rag-personalization-prompt} (top).
    \item RAG with Random User Profiles \cite{lampqa}: Similar to the previous method, but retrieval is performed on randomly sampled user profiles instead of the actual user profile. This baseline assesses the impact of using mismatched user information.
    \item PlanPers \cite{lampqa}: This method first retrieves information from the user profile using the question, then generates a high-level response plan based on the documents and the question. Conditioned on the plan, the retrieved documents, and the question, the LLM generates the final personalized response. This method uses the prompts shown in Figure~\ref{fig:planpers-prompt} for plan and response generation. We refer the reader to \citet{lampqa} for details.
    
    \item Offline RL RAG-Personalization: To compare with an approach that leverages scalar reward signals for training personalized LLMs, we implement Expectation-Maximization Offline Reinforcement Learning \cite{singh2024beyond, salemi2025reasoningenhancedselftraininglongformpersonalized}, using the downstream evaluation metric as the scalar reward. This algorithm is used for training due to the similarity of its training loop to that of our proposed method, making it a fair basis for comparison between learning from NLF and scalar reward. Similar to RAG-Personalization baseline, this method begins by retrieving a set of documents from the user profile. Based on the retrieved content, a set of $16$ candidate responses is generated for each question using the prompt shown in Figure~\ref{fig:rag-personalization-prompt} (top). The downstream task metric that scores outputs based on the question narrative and personalized rubrics is then applied to these responses, and the one with the highest score is selected to supervise the next iteration of training. The model is trained for three iterations under the same configuration as our method, with the best checkpoint from all iterations used for evaluation. This comparison enables an empirical assessment of the efficiency and effectiveness of NLF-based optimization versus scalar reward-based optimization.
\end{itemize}
All baselines are evaluated under the same setup and conditions as \ourmethod, including identical configurations for maximum input and output lengths, training budget, number of retrieved documents, retrieval model, and generation temperature.

\begin{figure*}
    \centering
    \includegraphics[width=\textwidth]{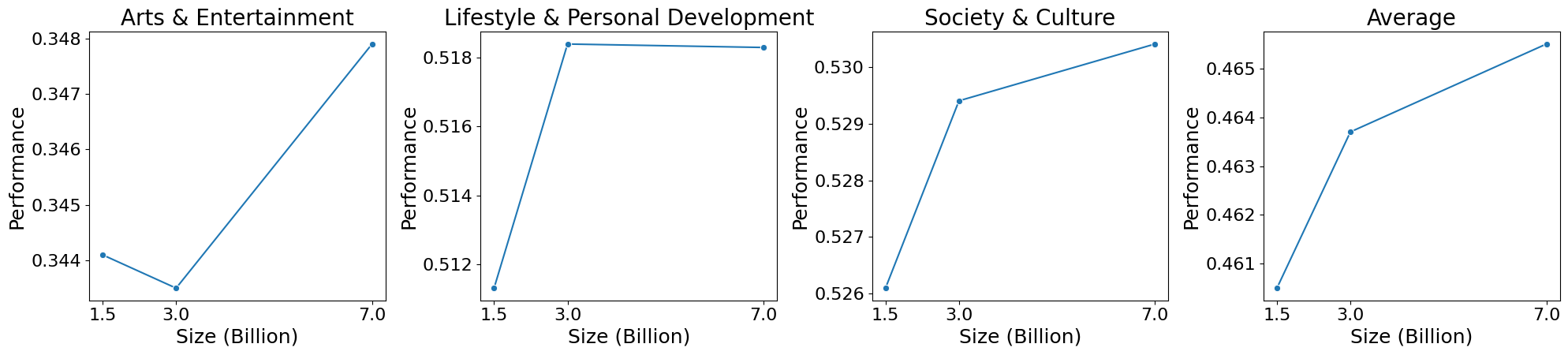}
    \vspace{-0.6cm}
    \caption{Effect of feedback model size on the performance of \ourmethod.}
    \label{fig:size-feedback-test}
\end{figure*}

\begin{figure*}
    \centering
    \includegraphics[width=\textwidth]{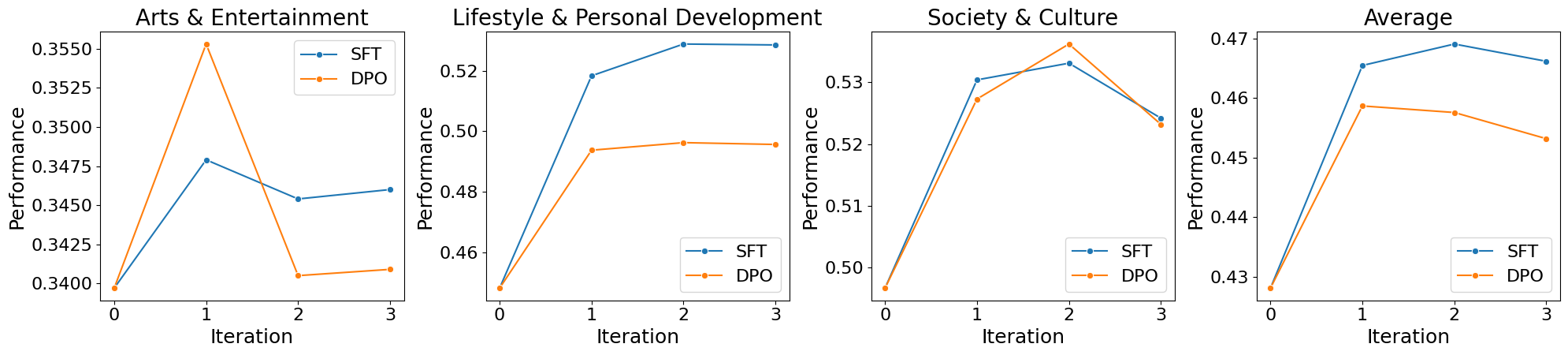}
    \vspace{-0.6cm}
    \caption{Effect of optimization method for policy model in \ourmethod in different training iterations.}
    \label{fig:iteration-optimization-test}
    % \vspace{-0.2cm}
\end{figure*}

\subsection{Empirical Results}

\subsubsection*{\textbf{Comparison with the Baselines:}}

The results of our method and the baselines on the LaMP-QA benchmark datasets are presented in Table~\ref{tab:main-results-test}. As shown, \ourmethod statistically significantly outperforms all baselines in terms of average performance across the benchmark. More specifically, \ourmethod achieves statistically significant improvements over all baselines in 2 out of the 3 personalized question answering tasks. The only task where \ourmethod does not outperform the baselines is Art \& Entertainment. These results highlight the effectiveness of learning from natural language feedback for improving personalization in response generation based on user preferences.

Table~\ref{tab:main-results-test} reports the runtime for each method. Among them, the non-personalized LLM yields the lowest runtime, primarily because it processes shorter inputs and incurs no retrieval overhead. In contrast, all RAG-based personalization methods---including \ourmethod---have higher runtime due to the added cost of retrieving relevant user profile documents. The highest runtime is observed for the PlanPers baseline \cite{lampqa}, which is nearly twice as slow as \ourmethod due to two step generation method used in this method, yet yields significantly lower performance. Overall, these results demonstrate that \ourmethod provides superior personalization performance with runtime costs comparable to the most efficient personalized baselines.

\subsubsection*{\textbf{Effect of Optimizing Feedback Model:}}

To examine the effect of training the feedback model on the performance of \ourmethod, we conduct two sets of experiments: one in which both the policy and feedback models are updated after each iteration, and another in which the feedback model remains frozen while only the policy model is trained. The results are reported in Figure~\ref{fig:iteration-test}. As shown, jointly training the feedback model to align with the evolving policy model consistently outperforms the frozen-feedback setup across all datasets. These findings highlight the importance of optimizing the feedback model in each iteration to match the updated capabilities of the policy model, thereby enabling the generation of more effective and informative feedback.

\subsubsection*{\textbf{Effect of Number of Training Iterations ($T$):}}

To investigate the impact of the number of training iterations ($T$) on the performance of \ourmethod, we train the model for up to three iterations and evaluate it after each iteration. The results in Figure~\ref{fig:iteration-test} show that the performance improves during the first two iterations but plateaus in the third. It also indicates that this plateau effect is more pronounced when the feedback model is untrained, highlighting the importance of optimizing the feedback model on the performance. These observations suggest that continued training with \ourmethod yields diminishing returns after a few iterations, and that joint optimization of both the policy and feedback models is crucial for maximizing effectiveness.

\subsubsection*{\textbf{Effect of Feedback Model Size:}}

To examine how the size and capabilities of the feedback model---which plays a central role in guiding the policy model during training---affect its effectiveness in helping the policy model to learn user preferences, we conduct experiments using instruction-tuned Qwen 2.5 of different parameter sizes: 1.5 billion, 3 billion, and 7 billion. These experiments are carried out over a single training iteration, and the results are presented in Figure~\ref{fig:size-feedback-test}. As shown, larger feedback models consistently lead to better performance of the policy model, suggesting that increased capacity enables the feedback model to generate more informative and actionable feedback. These findings underscore the importance of using strong feedback providers to more effectively supervise and guide the learning process of the policy model.

\begin{figure*}
    \centering
    \includegraphics[width=\textwidth]{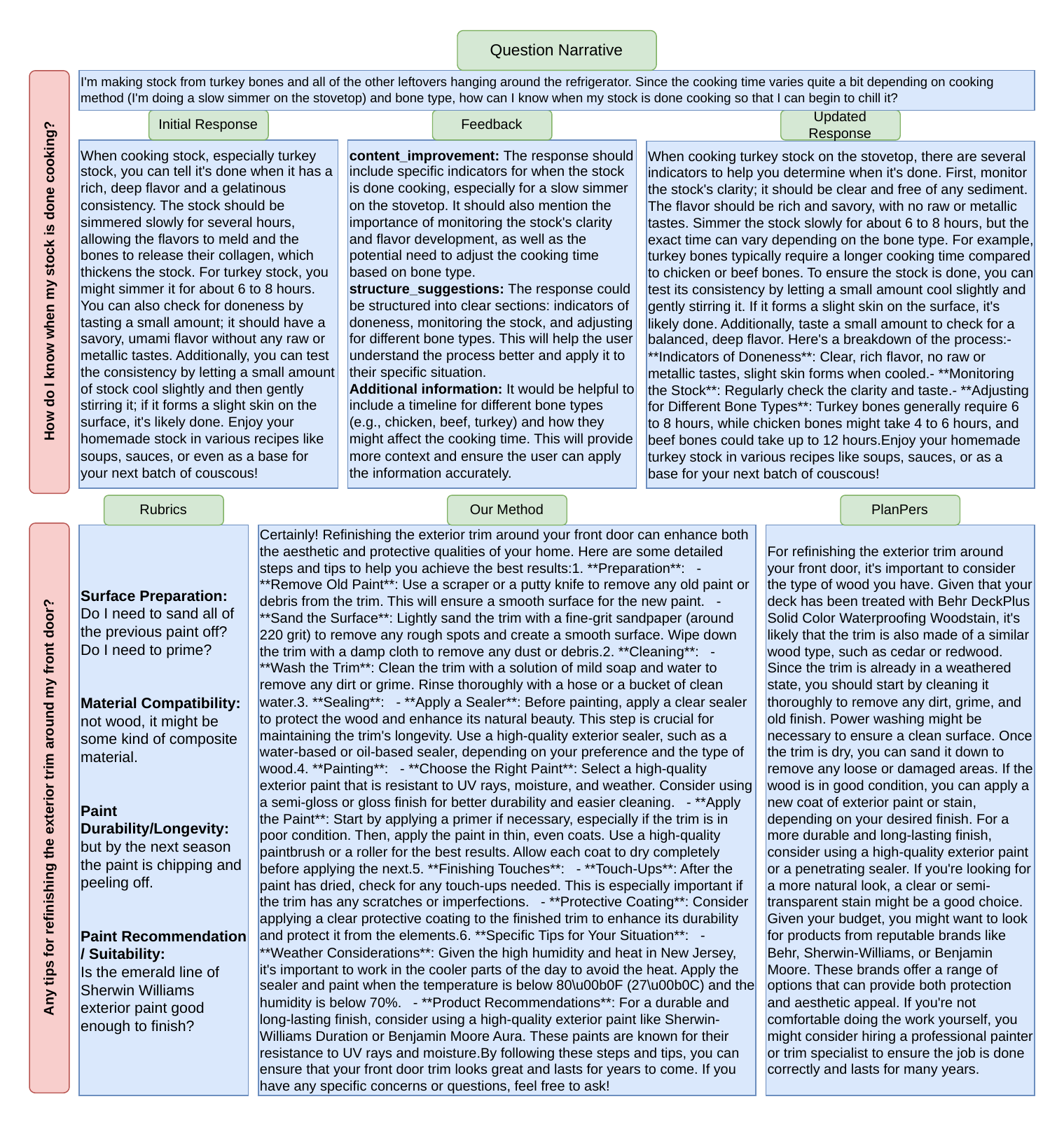}
    \vspace{-0.8cm}
    \caption{Case study illustrating the initial response, feedback, and updated response (top row), and a comparison between \ourmethod and the best-performing baseline, PlanPers \cite{lampqa} (bottom row).}
    % \vspace{-0.2cm}
    \label{fig:case-study}
\end{figure*}

\begin{figure}
    \centering
    % \vspace{-0.4cm}
    \includegraphics[width=0.8\linewidth]{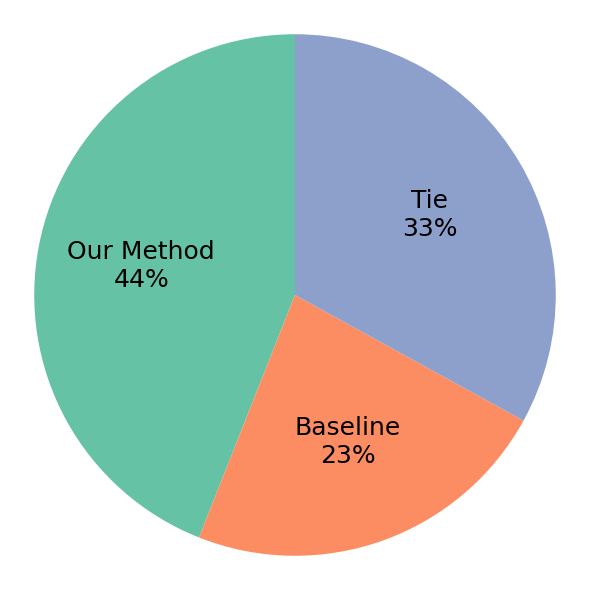}
    \vspace{-0.4cm}
    \caption{Results of human evaluation between \ourmethod and the best performing baseline, PlanPers \cite{lampqa}.}
    % \vspace{-0.4cm}
    \label{fig:human-eval}
\end{figure}

\subsubsection*{\textbf{Effect of Optimization Method for Policy Model:}}

As described in Section~\ref{sec:optimization}, once the updated output is generated using feedback, we train the policy model to reproduce this updated output via supervised fine-tuning (SFT) using a cross-entropy loss. An alternative approach is to optimize the model using methods such as Direct Preference Optimization (DPO) \cite{dpo}, which maximizes the likelihood of the updated output while minimizing the likelihood of the initial output. To investigate the effectiveness of this alternative, we perform a comparative experiment and report the results in Figure~\ref{fig:iteration-optimization-test}. The results show that, with the exception of the first iteration on the Art \& Entertainment dataset, SFT performs on par with or significantly better than DPO. We hypothesize that this outcome stems from a key assumption in DPO: it treats the initial and updated outputs as distinctly contrasting pairs. However, as training progresses, the policy model begins to generate outputs that are already close to the updated ones. In such cases, enforcing a strong separation between the two---as DPO does---may unnecessarily distort the model's output distribution. In contrast, SFT more naturally accommodates these subtler improvements, allowing the model to refine its outputs without overcorrecting.

\subsubsection*{\textbf{Human Evaluation:}}

We conduct a human evaluation to compare \ourmethod with the best-performing baseline, PlanPers \cite{lampqa}, from human perspective. We randomly sampled 100 examples from the LaMP-QA benchmark. Each example was independently evaluated by two human annotators, who were asked to compare the responses based on the rubric aspects and question narrative, and to determine which response better addressed the question—or if the responses were equally good. The inter-annotator agreement, measured using Cohen's kappa, is 0.7832, indicating a high level of consistency between annotators. The results of this evaluation are presented in Figure~\ref{fig:human-eval}. In 44\% of the cases, \ourmethod was preferred by the annotators for better addressing the rubric aspects. In 23\% of the cases, the PlanPers baseline was favored. The remaining 33\% were judged as ties. These demonstrate that \ourmethod produces responses that are more aligned with user-specific rubric aspects from human perspective, indicating its effectiveness in personalized question answering.

\subsection{Case Study}

This section presents case studies of initial responses, feedback, and updates during training, along with post-training outputs from the policy model compared to the top baseline.

\subsubsection*{\textbf{Initial response, feedback, and updated response during training:}}

As shown in Figure~\ref{fig:case-study} (top row), the initial response to the user's question about determining when turkey stock is done cooking was generally informative but lacked structure and omitted key contextual details from the user's narrative---particularly the focus on slow stovetop simmering and the influence of different bone types on cooking duration. The feedback addressed these shortcomings by recommending the inclusion of concrete doneness indicators (e.g., flavor, consistency, clarity), specific advice for stovetop cooking, and considerations for bone variability. It also suggested organizing the response into clear, distinct sections to enhance clarity. The updated response effectively incorporates these recommendations by outlining clear signs of doneness, offering practical monitoring strategies, and providing estimated cooking times based on bone type. Presented in a well-structured format, the revised response better reflects the user's context and yields a more personalized, informative, and user-aligned response.

\subsubsection*{\textbf{Comparing \ourmethod and the best-performing baseline after training:}}

As shown in Figure~\ref{fig:case-study} (bottom row), \ourmethod delivers a more tailored response than PlanPers by directly addressing the user's specific context and aligning more closely with the personalized evaluation rubrics. While PlanPers provides general guidance on refinishing exterior trim, \ourmethod goes further by incorporating details that reflect the user's concerns---such as previous issues with paint durability, uncertainty about the trim material, and questions about the suitability of Sherwin-Williams Emerald paint. The response includes specific advice on surface preparation, compatibility with composite materials, and environmental factors like local climate. This leads to higher scores across rubric dimensions including Material Compatibility, Paint Recommendation, and Durability/Longevity, demonstrating that \ourmethod more effectively captures user intent and provides context-aware, actionable guidance.

\section{Conclusions and Future Work}

% We introduced \ourmethod, a new framework for personalized response generation that replaces scalar rewards with natural language feedback as the primary supervision signal. By leveraging user-specific feedback grounded in both the user profile and question narrative, \ourmethod provides informative and actionable guidance for training personalized LLMs. Our iterative training loop---alternating between feedback generation and policy refinement---enables the policy model to internalize personalization strategies without requiring feedback at inference time. Experimental results on LaMP-QA showed that \ourmethod consistently outperforms existing personalized and non-personalized baselines and is also preferred by human evaluators.

% For future work, we plan to extend this feedback-based framework beyond response-level generation to include feedback over reasoning traces, enabling more personalized and transparent multi-step reasoning. Additionally, we aim to apply this method to a broader range of personalization tasks beyond question answering and investigate its effectiveness on different classes of LLMs, including reasoning-focused models. These directions will help assess the generality and adaptability of natural language feedback as a supervision mechanism for personalized text generation.

We introduced \ourmethod, a new framework for personalized response generation that replaces scalar rewards with natural language feedback as the primary supervision signal. By leveraging user-specific feedback grounded in both the user profile and question narrative, \ourmethod provides informative and actionable guidance for training personalized LLMs. In contrast to conventional reward modeling pipelines that compress rich user preferences into a single scalar signal, our approach preserves the structure and semantics of feedback, enabling more fine-grained alignment between model behavior and user intent. Our iterative training loop---alternating between feedback generation and policy refinement---enables the policy model to internalize personalization strategies without requiring feedback at inference time, effectively amortizing the cost of feedback into the training phase. This design not only improves sample efficiency but also allows the model to generalize personalization patterns across users and contexts. Experimental results on LaMP-QA showed that \ourmethod consistently outperforms existing personalized and non-personalized baselines and is also preferred by human evaluators, indicating that natural language feedback can serve as a scalable and expressive alternative to traditional supervision signals.

For future work, we plan to extend this feedback-based framework beyond response-level generation to include feedback over reasoning traces, enabling more personalized and transparent multi-step reasoning. Such an extension could allow the model to align not only its final outputs but also its intermediate decision-making process with user-specific preferences, potentially improving both interpretability and controllability. Additionally, we aim to apply this method to a broader range of personalization tasks beyond question answering, including dialogue systems, recommendation-oriented generation, and long-form content creation, where user intent is often more diffuse and evolving. We also plan to investigate its effectiveness on different classes of LLMs, including reasoning-focused models, as well as its interaction with emerging training paradigms such as tool use and agentic workflows. Finally, exploring the robustness of natural language feedback under noisy or partially observed user profiles, and understanding its role in mitigating biases in personalized systems, remain important open directions. These efforts will help assess the generality and adaptability of natural language feedback as a supervision mechanism for personalized text generation.

\section*{Acknowledgments}

This work was supported in part by the Center for Intelligent Information Retrieval, in part by an award from Adobe Systems, Inc., in part by NSF grant number \#2143434, and with support from Google.org. Any opinions, findings and conclusions or recommendations in this material are those of the authors and do not necessarily reflect those of the sponsors.

% TBD

\bibliographystyle{ACM-Reference-Format}
\bibliography{sample-base}

\end{document}